\documentclass{article}

\usepackage[preprint]{corl_2026} 
\usepackage{booktabs}
\usepackage{makecell}
\usepackage{wrapfig}
\usepackage[table]{xcolor}
\usepackage{xcolor}
\usepackage{amsmath}
\usepackage{graphicx}
\usepackage{amssymb}
\usepackage{multirow}
\usepackage{pifont}
\usepackage{siunitx}

\newcommand{\cmark}{\textcolor{green}{\ding{51}}}
\newcommand{\xmark}{\textcolor{red}{\ding{55}}}

\title{
LadderMan: Learning Humanoid\\Perceptive Ladder Climbing
}

\author{
  Siheng Zhao$^{1,2}$, Yuanhang Zhang$^{1}$, Ziqi Lu$^{1}$, Pieter Abbeel$^{1,3}$, Rocky Duan$^{1}$,\\ \textbf{Koushil Sreenath$^{1,3}$, Yue Wang$^{2}$, C. Karen Liu$^{1,4\dagger}$, Guanya Shi$^{1,5\dagger}$}\\
  $^{1}$Amazon FAR, $^{2}$USC, $^{3}$UC Berkeley, $^{4}$Stanford University, $^{5}$CMU, $^\dagger$Amazon FAR Co-Lead
}

\begin{document}
\maketitle
\vspace{-20pt}
\begin{figure}[htbp]
  \centering
  \includegraphics[width=\linewidth]{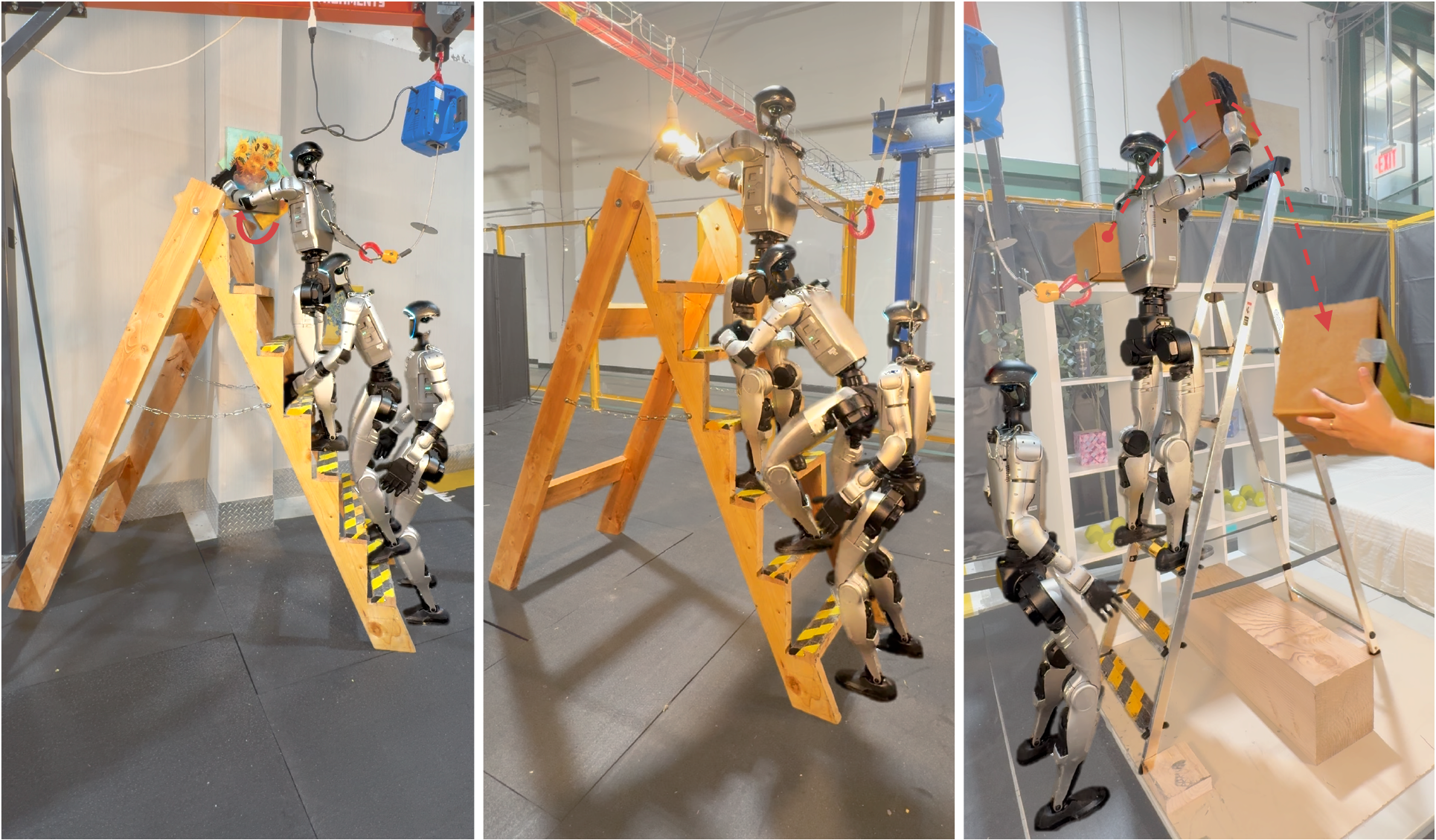}
  \caption{
  \textbf{LadderMan} enables a Unitree G1 humanoid to robustly climb diverse ladders using a single perceptive policy via zero-shot sim-to-real transfer. Building on the learned climbing capability, LadderMan further supports stable whole-body on-ladder manipulation via teleoperation, including adjusting paintings, replacing light bulbs, and box handover.
  }
  \label{fig:teaser}
  \vspace{-6pt}
\end{figure}

\begin{abstract}
Humanoid robots hold great promise for operating in human-centered environments, yet ladder climbing remains one of the most challenging tasks due to sparse footholds and handholds, complex whole-body coordination, and sensitivity to perception and control errors. We present \textbf{LadderMan}, a unified system that enables humanoid robots to robustly climb diverse ladders and perform manipulation under such constrained conditions. Our climbing policy is built on a scalable two-stage learning pipeline, where we use hybrid motion tracking to learn multiple climbing experts from a single reference motion, and distill these experts into a unified depth-based visuomotor climbing policy via hybrid imitation and reinforcement learning. To enable real-world deployment, we leverage vision foundation models to bridge the sim-to-real gap in depth perception. 
Building on the learned climbing policy, we further train a separate manipulation policy using a dual-agent formulation, allowing stable on-ladder manipulation via teleoperation.
Experiments demonstrate that LadderMan achieves robust ladder climbing across a wide range of geometries, successfully transfers to real-world hardware in a zero-shot manner, and supports various manipulation tasks under challenging ladder constraints. Video results are available on the \href{https://ladderman-robot.github.io/}{project website}.
\end{abstract}


\section{Introduction}
Enabling humanoid robots to robustly climb and operate on ladders is a key milestone toward deploying them in industrial environments such as construction sites, warehouses, and maintenance tasks. 
However, unlike walking on flat ground or stairs, ladder climbing requires precise and sustained coordination between upper and lower body contacts under sparse, discontinuous, and narrow footholds and handholds~\citep{HAMMER199221}. Small perception errors or control inaccuracies can easily lead to catastrophic failure. This challenge is further amplified in real-world environments, where ladders vary significantly in inclination, rung spacing, and structural design. 

Despite recent advances in humanoid locomotion over uneven terrain~\citep{zhang2026rpl,zhuang2024humanoid,wang2025beamdojo}, most existing approaches remain dominated by lower-body behaviors and do not fully exploit the whole-body capabilities of humanoids. Prior work on ladder climbing~\citep{4651212,6907259,Joris2016,8206364,8206565,8968175,xiaosun2021} largely relies on carefully engineered model-based control without incorporating perception into the control loop, and requires accurate environment modeling, task-specific tuning, and specialized hardware design.
These approaches typically exhibit slow climbing speeds, limited robustness, and poor generalization beyond specific ladder configurations. Moreover, they do not extend to on-ladder manipulation~\citep{weng2025hdmi,zhao2025resmimic}.

In this work, we ask: \textit{can we learn a visuomotor system that enables humanoid robots to robustly climb diverse ladders and perform on-ladder manipulation while maintaining stable balance, without additional hardware modifications?}

To this end, we propose \textbf{LadderMan}, a unified system for perceptive humanoid ladder climbing and on-ladder manipulation.
(1) Our climbing policy is built on a scalable two-stage learning pipeline. In the first stage, we introduce hybrid motion tracking to learn multiple expert policies from a single reference motion without requiring large motion datasets. To achieve this, we use relaxed upper-body tracking and incorporate ladder-centric contact tracking and climbing rewards. In the second stage, we distill these experts into a single depth-based visuomotor policy via hybrid imitation and reinforcement learning (RL), resulting in a unified policy that generalizes robustly across diverse ladder geometries.
(2) To enable real-world deployment on narrow and sparse ladder structures under background clutter, we leverage a vision foundation model~\citep{wen2026fastfoundationstereo} (VFM) to bridge the gap between simulated and real-world depth, instead of relying on extensive depth randomization and manual tuning.
(3) Building on the ladder climbing capability, we further investigate whole-body manipulation under ladder constraints. 
We adopt a dual-agent learning framework that decouples lower-body stabilization from upper-body manipulation~\citep{zhang2025falcon}, enabling the humanoid to perform manipulation tasks such as adjusting paintings and tightening light bulbs, while maintaining balance on ladder, which is difficult to achieve with off-the-shelf whole-body teleoperation policies~\citep{ze2025twist2,luo2025sonic}.

As shown in Fig.~\ref{fig:teaser}, our experiments demonstrate that LadderMan enables robust zero-shot sim-to-real ladder climbing across diverse ladder geometries, including variations in inclination, rung spacing, and structural design. The learned climbing policy achieves strong generalization in simulation, competitive climbing speeds compared to humans, and successful deployment on real-world humanoid hardware without additional hardware modifications. 
Furthermore, LadderMan supports stable on-ladder manipulation through a dual-agent teleoperation framework, outperforming off-the-shelf whole-body teleoperation policies under ladder constraints. 
All training and inference code, together with the deployable models, will be open-sourced.

\section{Related Works}
\textbf{Robot Ladder Climbing.}
Robot ladder climbing has long been considered a challenging problem in robotics, and was included as a key task in DARPA Robotics Challenge~\citep{darpa_drc}. A large body of prior work~\citep{4651212,6907259,Joris2016,8206364,8206565,8968175,xiaosun2021} has explored enabling humanoid robots to climb ladders using model-based control. These approaches typically assume access to accurate environment models and rely on task-specific tuning to ensure stable contacts. As a result, they are often limited to specific ladder configurations and do not incorporate perception as part of the control loop, making generalization to unseen ladders challenging.
More recently, RL has been explored for ladder climbing. \citet{vogel2025robustladder} demonstrates quadruped ladder climbing using an end-to-end policy, highlighting the potential of learning-based approaches for ladder traversal. However, compared with quadrupeds, humanoid ladder climbing introduces substantially greater challenges due to heterogeneous limbs, unstable support polygons, and the need for coordinated whole-body contacts between hands and feet. In addition, their method relies on privileged inputs such as motion capture and known ladder geometry, without incorporating perception directly into the control policy, and further requires task-specific hardware modifications.
In contrast, our work learns a perceptive humanoid ladder climbing policy that directly operates on depth observations and generalizes across diverse ladder geometries. Building on this capability, we further enable manipulation under ladder constraints.

\textbf{Perceptive Locomotion for Legged Robots.}
Perceptive locomotion enables legged robots to traverse complex terrains using exteroceptive sensing such as LiDAR and depth cameras. 
Early approaches typically rely on LiDAR-based mapping~\citep{wang2025beamdojo,scirobotics.abk2822,long2024learninghumanoidlocomotionperceptive,scirobotics.adi7566,scirobotics.adv3604,fankhauser2018probabilistic,ben2025gallant}, where raw sensor data are converted into terrain representations such as elevation maps to guide foothold selection and control. 
While effective for moderately structured environments, these methods depend heavily on accurate calibration, state estimation, and online mapping, which can introduce additional latency and sim-to-real gap. 
Furthermore, elevation maps struggle to capture thin and sparse structures such as ladder rungs with sufficient precision.
To overcome these limitations, more recent work has explored end-to-end visuomotor policies that directly use depth images~\citep{zhang2026rpl,zhuang2024humanoid,pmlr-v205-agarwal23a,cheng2023parkour,zhuang2023robot,yang2021learning} for locomotion control. 
These approaches have demonstrated impressive performance on challenging terrains including stairs~\citep{pmlr-v205-agarwal23a}, stepping stones~\citep{wang2025beamdojo}, and parkour-style environments~\cite{cheng2023parkour}. 
However, most existing methods rely heavily on reward shaping, which tends to produce lower-body-dominated behaviors when applied to humanoids~\citep{zhang2026rpl,zhuang2024humanoid,wang2025beamdojo,long2024learninghumanoidlocomotionperceptive,ben2025gallant}.
As a result, they do not fully exploit the whole-body coordination capabilities required for humanoid systems.
Concurrent work on perceptive whole-body humanoid locomotion~\citep{wu2026perceptivehumanoidparkourchaining,zhang2026learningwholebodyhumanoidlocomotion} demonstrates promising results on tasks such as obstacle climbing and vaulting which mainly focus on agility and dynamic interactions. In contrast, ladder climbing requires fine-grained and highly constrained multi-contact coordination on thin and sparse structures, placing significantly higher demands on perception and control accuracy.

\section{Method}
\begin{figure}[!t]
\vspace{-10pt}
\centering
\includegraphics[width=\linewidth]{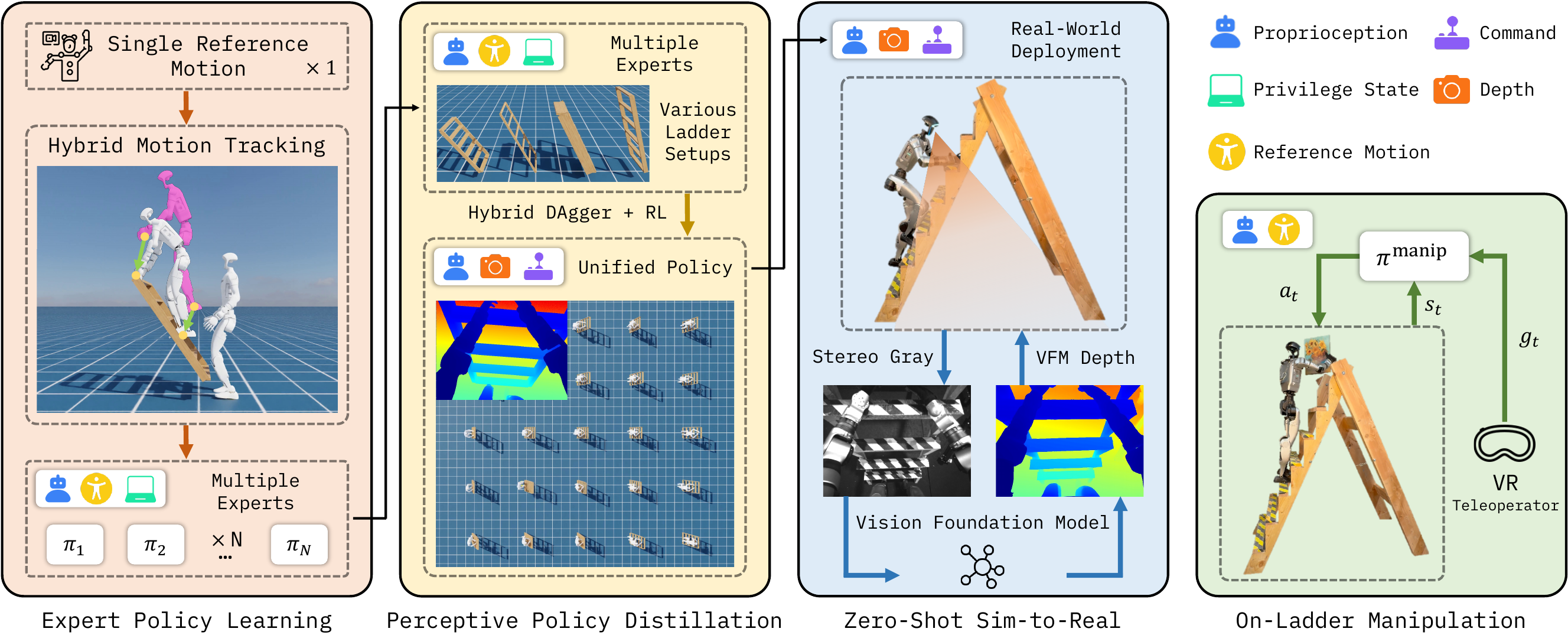}
\caption{\textbf{Overview of LadderMan.} Starting from a single reference motion, we first learn multiple expert climbing policies for different ladder geometries using hybrid motion tracking. These experts are then distilled into a unified visuomotor policy through hybrid DAgger and RL. Building on the climbing policy, we further train a dual-agent manipulation policy for stable on-ladder teleoperation. To bridge the sim-to-real perception gap, we use VFM for robust real-world deployment.
}
\label{fig:method}
\vspace{-10pt}
\end{figure}

Our goal is to learn a visuomotor system that enables humanoid robots to robustly climb diverse ladders and further perform manipulation under such constrained settings. This problem is particularly challenging due to four coupled factors: (1) ladder climbing requires precise whole-body coordination with tightly coupled hand-foot contacts, (2) ladder geometries vary significantly in inclination and rung spacing, making generalization difficult, (3) perception and control errors during sim-to-real on thin and sparse structures such as ladder rungs can easily accumulate and lead to catastrophic failures, and (4) manipulation under ladder constraints requires maintaining stable balance while executing upper-body interaction tasks.
To address these challenges, we design LadderMan as a system with four key components: learning multiple expert climbing policies from a single reference motion via hybrid motion tracking (Sec.~\ref{section:3.1}); distilling expert policies into a unified visuomotor climbing policy through hybrid imitation and RL (Sec.~\ref{section:3.2}); bridging the sim-to-real perception gap using a vision foundation model (Sec.~\ref{section:3.3}); enabling stable on-ladder manipulation through dual-agent learning (Sec.~\ref{section:3.4}).
An overview of LadderMan is shown in Fig.~\ref{fig:method}.

\subsection{Learning Multiple Expert Policies from Single Reference Motion}
\label{section:3.1}
Directly learning a single climbing policy that generalizes across diverse ladders is challenging. Therefore, in the first stage, we instead learn multiple state-based expert policies $\pi^{\text{expert}}_{\phi,z}$ corresponding to different ladder configurations $(\phi,z)$, where $\phi$ denotes ladder inclination and $z$ rung spacing. These expert policies are later distilled into a unified visuomotor policy $\pi^{\text{visual}}$. However, existing approaches are insufficient for humanoid ladder climbing: RL with reward shaping often leads to lower-body-dominated behaviors~\cite{zhang2026rpl}, while standard motion tracking methods~\cite{2018-TOG-deepMimic} require large motion datasets to cover diverse ladder geometries, which are difficult to obtain in practice.

To address these challenges, we introduce hybrid motion tracking, which combines relaxed upper-body tracking with task-driven adaptation to learn multiple expert climbing policies from a single reference. Details such as observation space and full reward terms are provided in Appendix~\ref{appendix:method1}.

\textbf{Reward.} Standard motion tracking~\cite{liao2025beyondmimic} reward is $r^{\text{tracking}}_X = \omega \exp(-\sigma \|\hat{X}[K] - X[K]\|_2),$
where $K$ denotes all tracked keypoints, $X$ denotes quantities such as joint positions and velocities, and $\hat{X}$ denotes the reference quantities. 
In hybrid motion tracking, we partition the body into upper-body keypoints $K_u$ and lower-body keypoints $K_l$, and use the following asymmetric tracking reward:
\begin{equation}\label{eq:tracking}
r^{\text{tracking}}_X =
\frac{\omega}{2} \exp\left(-\frac{\sigma}{2}\|\hat{X}[K_u] - X[K_u]\|_2\right)
+ \omega \exp\left(-\sigma \|\hat{X}[K_l] - X[K_l]\|_2\right).
\end{equation}
This asymmetric formulation encourages faithful reproduction of stable lower-body climbing behaviors while allowing the upper body to adapt contact locations according to ladder geometry.
To further guide adaptation, we introduce ladder-centric contact and climbing rewards:
\begin{equation}\label{eq:contact}
r^{\text{contact}} =\omega \hat{c} \exp\left(-\sigma \|P-\hat{P}\|_2 \right) + (1-\hat{c}), ~~r^{\text{task}} = \|R-\mathcal{G}\|_2,
\end{equation}
where $\hat{c}$ denotes the reference contact indicator, $P$ the current end-effector position, $\hat{P}$ the target contact position, $R$ the current root pose, and $\mathcal{G}$ the target root pose. 
The reference contact indicator is manually annotated once for the reference motion. The target contact position is automatically generated from the ladder geometry by assigning predefined contact locations on the target ladder rung for each hand and foot.
This hybrid formulation enables successful expert policy learning across diverse ladders from a single reference motion. 

\textbf{Observation.}
We extend the standard motion tracking observations (proprioception, reference motion, and privileged robot states) with relative poses between the humanoid pelvis and ladder.

\textbf{Termination and Domain Randomization.}
For termination, standard motion tracking terminates when either the local or global tracking error exceeds a predefined threshold. We further introduce a ladder-specific contact termination condition, where episodes terminate if required ladder contacts are lost for more than $30$ consecutive frames.
For domain randomization, in addition to physical parameter and initialization state randomization, we apply local perturbations to ladder configurations around each target configuration $(\phi,z)$, together with rung geometry and friction properties, to improve robustness during expert policy learning.

\subsection{Unified Climbing Policy via Hybrid Imitation and Reinforcement Learning}\label{section:3.2}

While individual expert policies $\pi_{\phi,z}^{\text{expert}}$ can handle specific ladder configurations $(\phi,z)$, practical deployment requires a unified policy that generalizes across diverse ladders and remains robust under compounding errors. A natural approach is to distill multiple experts into a single visuomotor policy $\pi^{\text{visual}}$ using DAgger-style imitation learning~\citep{dagger}. However, imitation alone is insufficient for this highly dynamic and contact-rich task: expert policies do not fully cover the configuration space, and small imitation errors can accumulate and lead to failure.
To address this issue, we combine DAgger with RL during policy distillation. The unified climbing policy is optimized with:
\begin{equation}
L(\pi^{\text{visual}}) = L_{\text{RL}}(\pi^{\text{visual}}) + \lambda D_{\text{KL}}
\left(\pi^{\text{visual}} \;\|\; \pi^{\text{expert}} \right),
\end{equation}
where $L_{\text{RL}}$ is the PPO~\citep{schulman2017proximalpolicyoptimizationalgorithms} objective using the same rewards as the expert policies, and the KL term encourages imitation of expert behaviors. The weight $\lambda$ is gradually annealed during training, allowing the policy to transition from imitation-driven initialization to task-driven optimization. Details such as observation space and model architecture are provided in Appendix~\ref{appendix:method2}.

\textbf{Observation and Depth Rendering.}
The unified policy $\pi^{\text{visual}}$ takes proprioception, depth observations, and a binary climbing direction command as input, and outputs whole-body actions. During training, depth images are rendered using NVIDIA Warp~\citep{macklin2022warp} for high-throughput simulation. The binary command specifies whether the humanoid should climb upward or downward.

\begin{figure}[!t]
\vspace{-10pt}
\centering
\includegraphics[width=\linewidth]{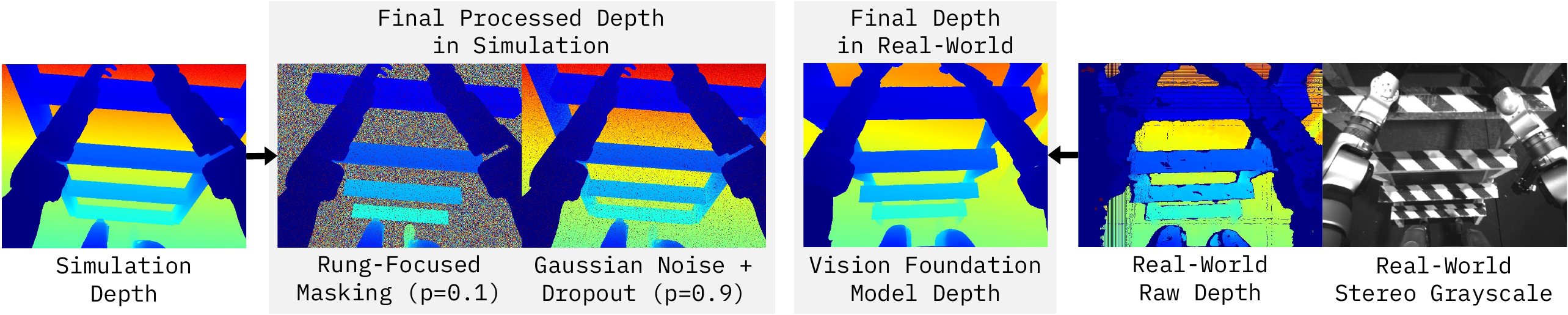}
\caption{We bridge the visual gap between simulation and real world by applying rung-focused masking, vision foundation model~\citep{wen2026fastfoundationstereo}, and minimalist noise augmentation.
}
\label{fig:depth_vis}
\vspace{-10pt}
\end{figure}

\subsection{Bridging Sim-to-Real Perception Gap with Vision Foundation Model}\label{section:3.3}
A major challenge in deploying perceptive humanoid policies is the sim-to-real gap in depth observations~\citep{zhu2026hikingwildscalableperceptive,zhuang2026deepwholebodyparkour}. For ladder climbing, this challenge is particularly severe due to the thin and sparse nature of ladder rungs, where depth inaccuracies can easily lead to contact failures. In practice, we observe two major sources of discrepancy between simulated and real-world depth observations: (1) \textit{unstructured noise} caused by sensor artifacts such as missing pixels and measurement noise, and (2) \textit{structured perception gap} arising from ladder-specific geometric variations that are difficult to model accurately and exhaustively in simulation. 

\textbf{Unstructured Noise.}
Prior works~\citep{zhang2026rpl,wu2026perceptivehumanoidparkourchaining,rudin2025parkourwild} typically address depth sim-to-real gap through extensive randomization, which often requires careful tuning of many hyperparameters. 
Instead, we use Fast-FoundationStereo~\citep{wen2026fastfoundationstereo}, a pretrained vision foundation model (VFM) for stereo depth estimation, to obtain real-world depth observations.  
As shown in Fig.~\ref{fig:depth_vis}, compared with raw depth sensing, VFM produces cleaner and more geometrically consistent depth observations with real-time inference speed, which is critical for humanoid ladder climbing. This substantially reduces the reliance on manually engineered depth randomization and improves robustness against sensor artifacts.
During training, we still apply lightweight depth randomization consisting of Gaussian noise $\mathcal{N}(0,0.005)$ and pixel dropout with probability $0.05$. The resulting depth values are clipped to the range $[0.1\,\mathrm{m},\,2\,\mathrm{m}]$.

\textbf{Structured Perception Gap.}
To improve robustness against ladder-specific geometric variations such as handrails and supports, we introduce rung-focused masking (RFM), which masks non-rung regions during training with probability $p=0.1$. As illustrated in Fig.~\ref{fig:depth_vis}, this encourages the policy to focus on task-relevant climbing structures rather than ladder-specific background geometry.

\subsection{On-Ladder Manipulation via Dual-Agent Learning}\label{section:3.4}
Building on the learned climbing policy, we further enable manipulation under ladder constraints. This setting is challenging because the humanoid must maintain stable balance at the top of the ladder while executing manipulation tasks. 
Directly applying an off-the-shelf teleoperation policy often leads to unstable whole-body behaviors during policy transition and manipulation. 

To address this issue, we train a separate manipulation policy
$\pi^{\text{manip}}(s_t,g_t) = \left[ \pi^{u}(s_t,g_t); \pi^{l}(s_t) \right]$, using a dual-agent formulation~\citep{zhang2025falcon}. 
During training, the two agents use separate actor and critic networks.
The lower-body agent $\pi^{{l}}$ is responsible for maintaining stable balance, while the upper-body agent $\pi^{{u}}$ tracks teleoperated manipulation targets. This decomposition decouples stabilization from manipulation, reducing interference between the two objectives and improving manipulation stability under ladder constraints. Refer to Appendix~\ref{appendix:method4} for more details.

\textbf{Observation, Action, and Reward.}
Both agents receive whole-body proprioception $s_t$ as input. $\pi^u$ additionally receives teleoperated upper-body motion targets $g_t$, including arm, wrist, and waist poses. During training, upper-body target joint poses are randomly sampled from the AMASS~\citep{AMASS} dataset, while during deployment they are generated through IK from teleoperation. The final action is formed by concatenating the outputs of the two agents: $a_t = [a_t^{{l}};a_t^{{u}}]$. $\pi^{{u}}$ is trained using the upper-body tracking reward $r_X^{\text{tracking}, u}$ defined in Eq.~\ref{eq:tracking}. $\pi^{{l}}$ is trained using the contact reward $r^{\text{contact}}$ and stabilization reward $r^{\text{task}}$ from Eq.~\ref{eq:contact}, which encourage stable ladder contacts and maintain the pelvis near the target on-ladder pose.
To increase manipulation workspace during training, upper-body collisions with the ladder are disabled in simulation.

\begin{figure}[!t]
\vspace{-10pt}
\centering
\includegraphics[width=\linewidth]{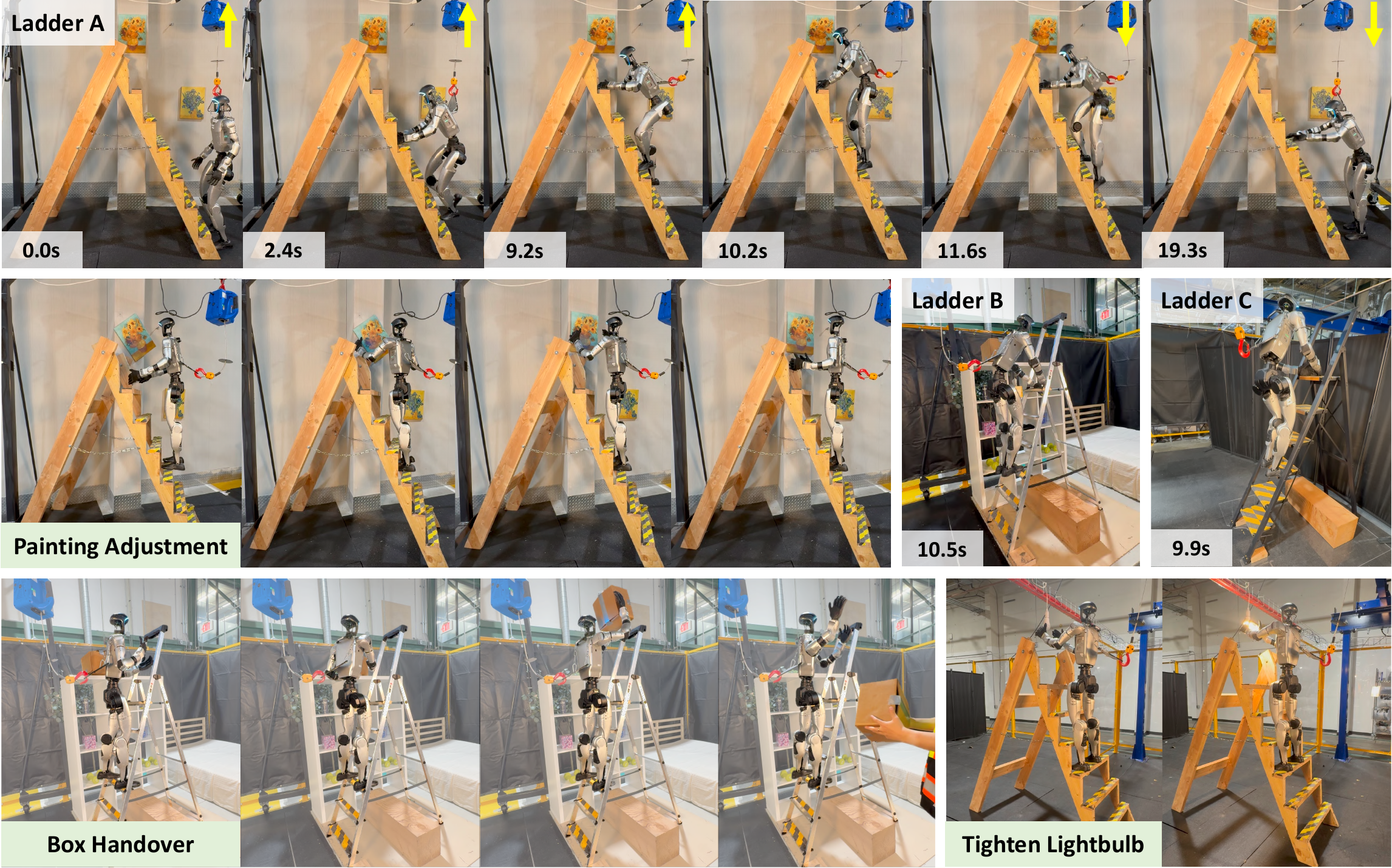}
\caption{
Real-world results of \textbf{LadderMan}. 
The humanoid performs zero-shot sim-to-real ladder climbing across diverse real-world ladders (A, B, and C) and executes various on-ladder manipulation tasks while maintaining stable balance.
Additional video results are available on our \href{https://ladderman-robot.github.io/}{website}.
}
\label{fig:real_result}
\vspace{-10pt}
\end{figure}

\section{Experimental Results}
We evaluate LadderMan through extensive simulation and real-world experiments to answer:

\textbf{Q1:} Can LadderMan robustly perform climbing and manipulation across diverse ladders?
\\[0\baselineskip]
\textbf{Q2:} Does the learned dual-agent policy improve manipulation stability under ladder constraints?
\\[0\baselineskip]
\textbf{Q3:}
Can hybrid motion tracking learn expert policies across diverse ladders from single reference?
\\[0\baselineskip]
\textbf{Q4:} Which system components are critical for robust sim-to-real transfer?

We use IsaacSim~\citep{NVIDIA_Isaac_Sim} for policy training and evaluation. All policies are deployed on a $1.3\,\mathrm{m}$ 29-DoF Unitree G1 humanoid without hardware modifications, using an onboard Intel RealSense D435i camera for perception and an onboard NVIDIA Jetson Orin for policy inference.
We use only a single reference motion for expert policy learning, collected using an OptiTrack motion capture system on a ladder with inclination angle $\theta=65.5^\circ$ and rung spacing $s=24.8\,\mathrm{cm}$. Additional experimental details are provided in Appendix~\ref{appendix:exp}.

\subsection{Real-World Evaluation}
\begin{figure}[!t]
\vspace{-10pt}
\centering
\includegraphics[width=\linewidth]{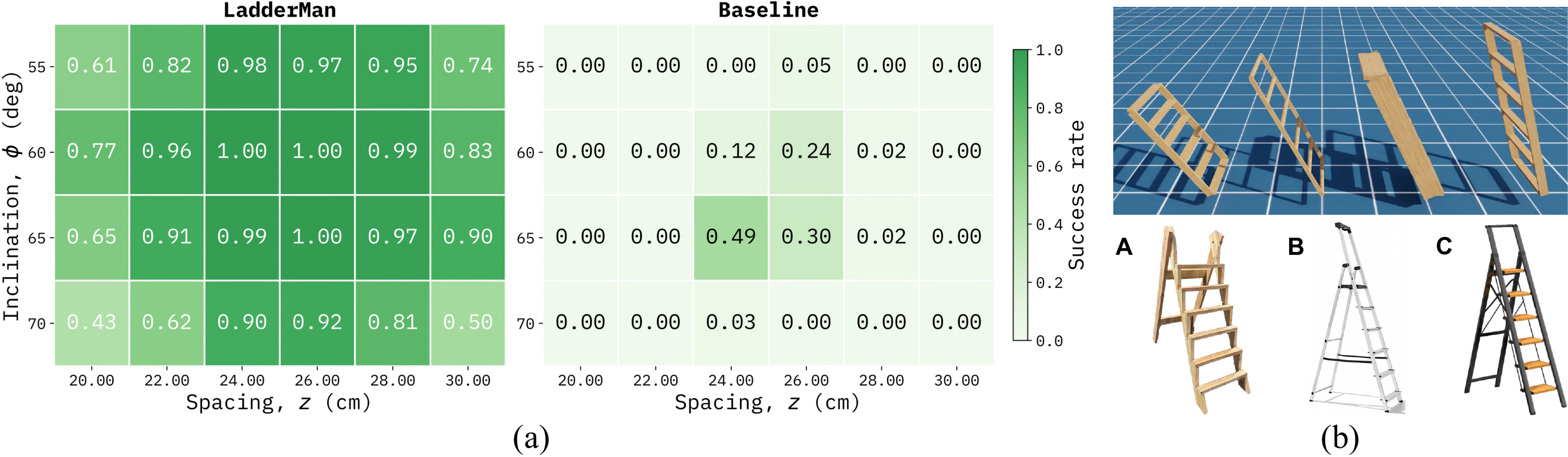}
\caption{(a) Success rates of LadderMan and the blind motion tracking baseline across ladder configurations with varying inclination angles and rung spacings. (b) Representative ladder geometries used in simulation \textit{(top)} and real-world evaluation \textit{(bottom)}. 
}
\label{fig:result_ladder}
\vspace{-10pt}
\end{figure}

\textbf{Experiment Setup.}
To answer \textbf{Q1}, we evaluate our climbing policy on three real-world ladders, including one custom-built ladder and two commercially available ladders with diverse geometries and materials, as shown in Fig.~\ref{fig:result_ladder}(b). We parameterize each ladder configuration as a tuple $(\phi, z, l, w)$, where $\phi$ denotes inclination angle, $z$ rung spacing, and $l,w$ the rung dimensions. The evaluated ladders correspond to:
$\text{A}$ $=$ $(65.5^\circ,24.8\mathrm{cm},60\mathrm{cm},8\mathrm{cm}),$
$\text{B}$ $=$ $(69.2^\circ,23.1\mathrm{cm},28\mathrm{cm},8\mathrm{cm}),$
$\text{C}$ $=$ $(64.8^\circ,25.5\mathrm{cm},30\mathrm{cm},20\mathrm{cm}).$ The reference motion is collected on Ladder A.

\textbf{Ladder Climbing Results.}
As shown in Fig.~\ref{fig:real_result} and Tab.~\ref{tab:ablation_sim2real}, LadderMan successfully climbs both upward and downward on Ladder A within $20$ seconds in a zero-shot sim-to-real setting.
We further demonstrate successful transfer across additional real-world ladders (Ladders B and C) with different geometries and materials. Despite significant variations across ladders, the learned policy maintains stable whole-body coordination and reliable multi-contact transitions throughout the process.

\begin{wrapfigure}{r}{0.5\textwidth}
\vspace{-10pt}
\centering
\includegraphics[width=1.\linewidth]{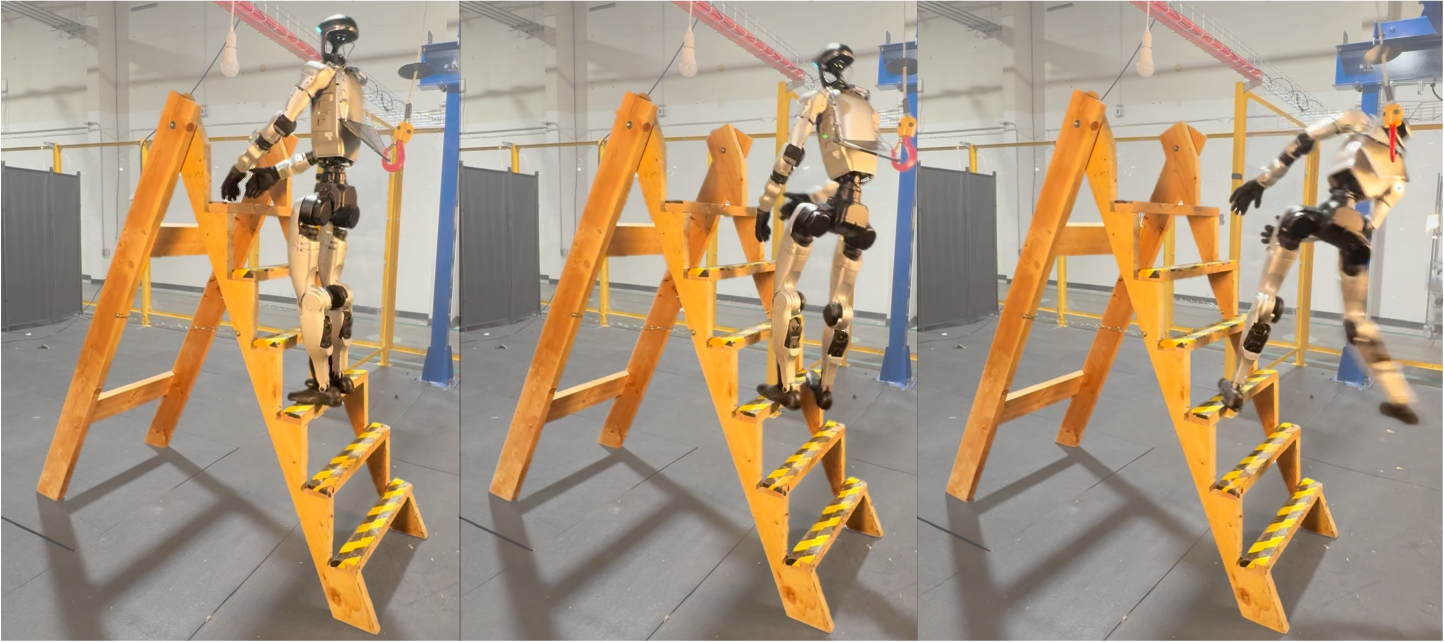}
\caption{Failure case of the off-the-shelf teleoperation policy TWIST2 under ladder constraints.
}
\vspace{-10pt}
\label{fig:twist2}
\end{wrapfigure}
\textbf{On-Ladder Manipulation Results.}
Building on the learned climbing policy, we further evaluate on-ladder manipulation in the real world to answer \textbf{Q1}. 
We use a PICO 4 Ultra VR headset for teleoperation, which provides whole-body motion streaming through a head-mounted display, handheld controllers, and two ankle-mounted trackers without requiring external motion capture systems.
As shown in Fig.~\ref{fig:real_result}, the humanoid performs diverse on-ladder manipulation tasks, including adjusting painting, replacing light bulb, and box handover, while maintaining stable balance.

To answer \textbf{Q2}, we compare the learned dual-agent policy against a state-of-the-art whole-body teleoperation policy TWIST2~\citep{ze2025twist2}. After reaching the top of the ladder using the climbing policy, control is switched to the teleoperation policy. 
As shown in Fig.~\ref{fig:twist2}, TWIST2 frequently loses balance, while our dual-agent policy maintains stable balance throughout manipulation due to decoupling lower-body stabilization from upper-body manipulation.

\textbf{Human Comparison.}
To evaluate practical performance, we compare the climbing speed of LadderMan against human measured for one male and one female subject across three different ladders. Humans achieve average climbing speed of $\sim 3.2\,\mathrm{s/rung}$, while LadderMan achieves a comparable speed of $\sim 3.4\,\mathrm{s/rung}$. Side-by-side comparison is provided in Appendix~\ref{appendix:exp_results}.

\subsection{Simulation Evaluation}
\textbf{Experiment Setup.} 
To further answer \textbf{Q1} quantitatively, we evaluate the learned climbing policy across ladders with varying rung spacing $z$ and inclination angle $\phi$. Specifically, we evaluate the policy on a discrete grid of ladder configurations with inclination angles $\phi \in \{55^\circ,60^\circ,65^\circ,70^\circ\}$ and rung spacings $z \in \{20,22,24,26,28,30\}\,\mathrm{cm}$, covering a broad range of real-world ladder geometries. For each ladder configuration, we further randomize rung size and shape (cuboid or cylindrical). Representative ladder geometries used in simulation are shown in Fig.~\ref{fig:result_ladder}(b). We report success rates over 100 evaluation episodes for each configuration.

\textbf{Baseline.} 
We compare the climbing policy of LadderMan against a blind motion tracking baseline trained without perceptual observations following the BeyondMimic~\citep{liao2025beyondmimic} training setup.

As shown in Fig.~\ref{fig:result_ladder}(a), LadderMan demonstrates strong generalization across a wide range of ladder configurations, achieving over 95\% success rate for ladders with spacing between $24\mathrm{cm}$ and $28\mathrm{cm}$ and inclination angles between $55^\circ$ and $65^\circ$. Even under more extreme configurations near the boundary of the evaluation space, the policy still maintains an average success rate of $\sim 57$\%.
In contrast, the blind motion tracking baseline achieves a peak success rate of only 49\% near the reference configuration and rapidly fails as ladder geometry deviates from the training setup, yielding near-zero success rates for most unseen configurations.

\subsection{Ablation Studies}
\begin{table}[t]
\vspace{-10pt}
\centering
\scriptsize
\begin{minipage}{0.58\linewidth}
\centering
\setlength{\tabcolsep}{3pt}
\begin{tabular}{l|ccc}
\toprule
\textbf{Method} & $(55^\circ, 20\si{\centi\meter})$ & $~~(65.5^\circ, 24.8\si{\centi\meter})^\star$ & $(70^\circ, 30\si{\centi\meter})$ \\
\midrule
DeepMimic~\citep{2018-TOG-deepMimic} & \xmark &  \cmark & \xmark  \\
\rowcolor{gray!20}
Ours & \cmark & \cmark & \cmark \\
 ~~~w/o contact tracking & \xmark & \cmark & \xmark \\
 ~~~w/o climb reward & \cmark & \cmark  & \xmark \\
\bottomrule
\end{tabular}
\vspace{6pt}
\caption{Ablation study on expert policy learning methods across different ladder geometries.~$^\star$ denotes the ladder configuration where the reference motion was collected.}
\label{tab:ablation_expert}
\end{minipage}
\hfill
\begin{minipage}{0.36\linewidth}
\centering
\setlength{\tabcolsep}{3pt}
\begin{tabular}{l|ccc}
\toprule
\textbf{Method} & Ladder A & Ladder B & Ladder C  \\
\midrule
\rowcolor{gray!20}
Ours & \textbf{9/10} & \textbf{6/10} & \textbf{7/10} \\
~~~w/o RL &  2/10 & 0/10 & 0/10 \\
~~~w/o VFM & 3/10 & 0/10 & 2/10 \\
~~~w/o RFM & 0/10 & 0/10 & 0/10  \\
\bottomrule
\end{tabular}
\vspace{6pt}
\caption{Real-world ablation of key components for robust sim-to-real transfer across $3$ different ladders.
}
\label{tab:ablation_sim2real}
\end{minipage}
\vspace{-16pt}
\end{table}
\textbf{Hybrid Motion Tracking Enables Expert Policy Learning Across Diverse Ladder Geometries.}
To answer \textbf{Q3}, we ablate hybrid motion tracking during expert policy learning.
We compare against: (1) standard DeepMimic-style motion tracking, (2) hybrid motion tracking without contact tracking, and (3) hybrid motion tracking without task rewards.
We use only a single reference motion captured on a ladder with $(\phi,z)=(65.5^\circ,24.8\mathrm{cm})$. For each method, we independently train expert policies on three representative ladder configurations:
$(55^\circ, 20\mathrm{cm}), (65.5^\circ, 24.8\mathrm{cm}), (70^\circ, 30\mathrm{cm})$
corresponding to ladders with smaller, identical, and larger geometry relative to the reference setup.
After training for $8,000$ iterations, we evaluate whether each method can successfully learn a stable climbing policy for the target ladder configuration. The evaluation metric is binary, indicating whether successful climbing behavior emerges during training.
As shown in Tab.~\ref{tab:ablation_expert}, standard motion tracking only succeeds when training on the reference ladder geometry and fails to learn stable climbing behaviors on substantially different configurations. In contrast, hybrid motion tracking successfully learns climbing behaviors across all evaluated ladder geometries from a single reference motion. Removing either contact tracking or task rewards degrades learning performance, highlighting the importance of task-driven adaptation for robust whole-body climbing skill acquisition.

\textbf{Key Components for Robust Sim-to-Real Transfer.} 
To answer \textbf{Q4}, we ablate several key system components critical for real-world deployment: (1) hybrid DAgger and RL, (2) vision foundation model (VFM)-based depth perception, and (3) rung-focused masking (RFM). We evaluate all variants in the real world over 10 trials with randomized initial poses.
As shown in Tab.~\ref{tab:ablation_sim2real}, incorporating RL substantially improves robustness and recovery behaviors, resulting in significantly higher real-world success rates than pure DAgger. In addition, both VFM-based perception and rung-focused masking contribute significantly to sim-to-real perception robustness and generalization. In particular, masking task-irrelevant ladder structures improves perception consistency and reduces failures caused by unseen ladder rail geometries that are difficult to model accurately in simulation.

\section{Conclusion and Limitation}
In this work, we present LadderMan, a unified perceptive system for humanoid ladder climbing and on-ladder manipulation. We propose a scalable two-stage learning pipeline for learning a unified visuomotor climbing policy. To enable zero-shot real-world deployment, we leverage vision foundation models to bridge the sim-to-real gap in depth perception.  Our results demonstrate robust climbing performance across diverse ladder geometries and stable on-ladder manipulation through a dual-agent teleoperation framework.

Despite these promising results, several limitations remain and open directions for future research. (1) Our climbing policy is primarily evaluated on inclined ladders up to $75^\circ$ and does not yet extend to vertical ladders, which require larger contact forces, tighter stability margins, and potentially different climbing strategies. (2) Our manipulation capability is currently limited by the absence of dexterous end-effectors. Extending the system with dexterous hands is an important direction toward more practical real-world applications.


\clearpage


\bibliography{example}  

\clearpage
\appendix

\section{Method}\label{appendix:method}

\subsection{Learning Multiple Expert Policies from a Single Reference Motion}
\label{appendix:method1}

In this stage, our goal is to learn state-based expert policies $\pi^{\text{expert}}_{\phi,z}$ corresponding to different ladder configurations $(\phi,z)$, where $\phi$ denotes the ladder inclination and $z$ the rung spacing.
We formulate the problem as a goal-conditioned reinforcement learning problem within a Markov Decision Process (MDP)
\begin{equation}
\mathcal{M}
=
\langle
\mathcal{O},
\mathcal{A},
\mathcal{T},
\mathcal{R},
\gamma
\rangle,
\end{equation}
where $\mathcal{O}$ is the observation space, $\mathcal{A}$ the action space, $\mathcal{T}$ the transition dynamics, $\mathcal{R}$ the reward function, and $\gamma$ the discount factor.
At time step $t$, the policy $\pi$ takes observation $o_t\in\mathcal{O}$ as input and outputs action $a_t$, which specifies target joint positions executed through a PD controller. The reward is defined as
\begin{equation}
r_t=\mathcal{R}(o_t,a_t),
\end{equation}
and the training objective is to maximize the expected cumulative discounted reward:
\begin{equation}
\mathbb{E}
\left[
\sum_{t=1}^{T}
\gamma^{t-1}r_t
\right].
\end{equation}

\textbf{Reference Motion Collection.}
Hybrid motion tracking requires only a single reference motion. We collect this motion using an OptiTrack motion capture system on a ladder with inclination angle $\phi=65.5^\circ$ and rung spacing $z=24.8\,\mathrm{cm}$. The reference motion consists of a complete climbing sequence and is used for learning all expert policies across different ladder configurations.
In addition to the motion trajectory, we manually annotate contact events for the four end-effectors (both hands and feet) once on the reference motion. These annotations are used to construct the reference contact indicator $\hat{c}$ and the ladder-centric contact targets used in the contact tracking reward.
Despite using only a single reference motion, our hybrid motion tracking formulation enables successful expert policy learning across a wide range of ladder geometries.

\begin{table}[b]
\scriptsize
\centering
\begin{tabular}{@{}c@{\hspace{0.0em}}c@{}}
\toprule
\begin{tabular}[t]{lcc}
\textbf{Observation Group}  & \textbf{Description} & \textbf{Equation}  \\
\midrule
\multirow{3}{*}{Reference Motion}
& reference body local pose & $\hat{p}_t$ \\
& reference joint position & $\hat{q}_t$ \\
& reference motion phase & $t/T$ \\
\midrule
\multirow{4}{*}{Proprioception}
& root angle velocity & $w_t$  \\
& root orientation & $\theta_t$ \\
& joint position & $q_{t-8:t}$  \\
& previous actions & $a_{t-3:t-1}$  \\
\bottomrule
\end{tabular}
&
\begin{tabular}[t]{|lcc}
\textbf{Observation Group}  & \textbf{Description} & \textbf{Equation}  \\
\midrule
\multirow{7}{*}{Privilege State}
& ladder relative pose & $p_t^{\text{root}} - p_t^{\text{ladder}}$ \\[0.47pt]
& body local pose & ${p}_t$ \\[0.47pt]
& body velocity & $\dot{p}_t$ \\[0.47pt]
& root velocity & $v_t$ \\[0.47pt]
& body pose error & $p_t- \hat{p}_t$ \\[0.47pt]
& body velocity error & $\dot{p}_t- \hat{\dot{p}}_t$ \\[0.47pt]
& joint position error & $q_t- \hat{q}_t$ \\[0.47pt]
\bottomrule
\end{tabular}
\end{tabular}
\vspace{6pt}
\caption{Observation space of the expert climbing policy.}
\label{tab:appendix_observation}
\end{table}
\textbf{Observation Space and Reward.}
The expert policy $\pi^{\text{expert}}_{\phi,z}$ takes (i) reference motion, (ii) proprioception, and (iii) privileged state information as observations, as shown in Table~\ref{tab:appendix_observation}.
The reward consists of three categories: motion tracking, contact tracking, and regularization rewards, as summarized in Table~\ref{tab:appendix_reward}. Motion tracking rewards follow the exponential form
\begin{equation}
r
=
\omega
\exp
\left(
-\frac{E}{\tau}
\right),
\end{equation}
where $\omega$ and $\tau$ denote the reward weight and temperature, respectively.
Specifically, the contact tracking reward is defined as
\begin{equation}
r^{\text{contact}}
=
\omega \hat{c}
\exp
\left(
-\frac{\|P-\hat{P}\|_2}{\tau}
\right)
+
(1-\hat{c}),
\end{equation}
where $\hat{c}$ denotes the reference contact indicator, $P$ the current end-effector position, and $\hat{P}$ the target contact position.
The reference contact indicator is manually annotated once for the reference motion. The target contact position is automatically generated from the ladder geometry by assigning predefined contact locations on the target ladder rung for each hand and foot.

\begin{table}[h!]
\small
\centering
\begin{tabular}{@{}c@{\hspace{0.0em}}c@{}}
\toprule
\begin{tabular}[t]{lccc}
\multicolumn{2}{l}{\textbf{Reward}} & \textbf{Weight} & \textbf{Temperature} \\
\midrule
\multicolumn{4}{l}{\textit{Motion Tracking}} \\
\multirow{2}{*}{Body Local Pose}
    & Upper Body & $1.0$ & $1.0$ \\
    & Lower Body & $2.0$ & $0.5$ \\
\multirow{2}{*}{Body Velocity}
    & Upper Body & $0.5$ & $5.0$ \\
    & Lower Body & $1.0$ & $2.5$ \\
\multirow{2}{*}{Joint Tracking}
    & Upper Body & $0.5$ &  $0.5$ \\
    & Lower Body & $1.0$ & $0.25$ \\
{Root Pose} & - & $1.0$ & $0.5$ \\[0.72pt]
\midrule
\multicolumn{4}{l}{\textit{Contact Tracking}} \\
\multicolumn{2}{l}{{Hands Contact Position}}     & $0.3$ & $0.3$ \\
\multicolumn{2}{l}{{Feet Contact Position}}    & $0.3$ & $0.3$ \\
\bottomrule
\end{tabular}
&
\begin{tabular}[t]{|lc}
\textbf{Reward} & \textbf{Weight} \\
\midrule
\multicolumn{2}{|l}{\textit{Regularization Penalties}} \\
Action Rate L2              & $0.1$ \\
Joint Position Limits L1    & $10.0$ \\
Joint Velocity Penalty      & $5e^{-4}$ \\
Joint Torque Limits L1      & $0.01$ \\
Feet Impact Force L2        & $1.0$ \\
Feet Slip                   & $0.5$ \\
Feet Air Time               & $5.0$ \\
\midrule
& \\
& \\
& \\
\bottomrule
\end{tabular}
\end{tabular}
\vspace{6pt}
\caption{Reward terms used for expert policy learning.}
\label{tab:appendix_reward}
\end{table}

\textbf{Domain Randomization and Termination.} As shown in Tab.~\ref{tab:appendix_randomization}, in addition to physical parameters and reference-state randomization, we further randomize ladder configurations by perturbing both ladder inclination $\phi$ and rung spacing $z$.
Training episodes terminate under three conditions: (i) root global pose error exceeds $0.5\,\mathrm{m}$ or $1.2\,\mathrm{rad}$ for 25 consecutive simulation steps; (ii) body local pose error exceeds $0.5\,\mathrm{m}$ or $1.2\,\mathrm{rad}$ for 25 consecutive simulation steps; (iii) required ladder contacts are lost for more than 30 consecutive simulation steps.
\begin{table}[t]
\vspace{-6pt}
\scriptsize
\centering
\begin{tabular}[t]{lc|lc}
\toprule
\textbf{Domain Randomization}  & \textbf{Sampling Distribution} & \textbf{Domain Randomization}  & \textbf{Sampling Distribution} \\
\midrule
\textit{Physical parameters} & & \textit{Reference initialization} & \\
Link mass & $\text{mass} \sim \mathcal{U} [0.9, 1.1] \times \text{mass}$ & Reference root position [m] & $\Delta x y z \sim \mathcal{U} [-0.01, 0.01]$\\
Torso COM offset [m] & $\Delta x y z \sim \mathcal{U} [-0.02, 0.02]$ & Reference root rotation [rad] & $\Delta r p y \sim \mathcal{U} [-0.05, 0.05]$ \\
Joint position offset [rad] & $\Delta q \sim \mathcal{U} [-0.01, 0.01]$ & Reference joint position [rad] & $\Delta q \sim \mathcal{U} [-0.05, 0.05]$ \\
Static friction coefficient & $\mu_{\text{static}} \sim \mathcal{U} [0.3, 7.2]$ & \textit{Ladder Configuration} & \\
Dynamic friction coefficient & $\mu_{\text{dynamic}} \sim \mathcal{U} [0.3, 3.6]$ & Ladder inclination [$^\circ$] & $\Delta \phi \sim \mathcal{U} [-1.0, 1.0]$ \\
Restitution coefficient & $e_{\text{rest}} \sim \mathcal{U} [0.0, 0.2]$ & Ladder rung spacing [m] & $\Delta z \sim \mathcal{U} [-0.01, 0.01]$ \\
\bottomrule
\end{tabular}
\vspace{6pt}
\caption{Domain randomization parameters used during expert policy training.}
\vspace{-6pt}
\label{tab:appendix_randomization}
\end{table}

\textbf{Training Details.}
We use PPO~\citep{schulman2017proximalpolicyoptimizationalgorithms} to train the expert policies. Training is performed in IsaacSim~\citep{NVIDIA_Isaac_Sim} with 4096 parallel environments on a single NVIDIA L40S GPU for 8,000 epochs. The actor and critic networks use MLP architectures of
$[512,256,256]$ and $[512,256,128]$, respectively.

\subsection{Unified Climbing Policy via Hybrid Imitation and Reinforcement Learning}
\label{appendix:method2}

In this stage, our goal is to learn a unified visuomotor climbing policy $\pi^{\text{visual}}$ that generalizes across diverse ladder geometries. The policy takes climbing commands, proprioception (as defined in Tab.~\ref{tab:appendix_observation}), and depth observations as input.

For depth rendering, we use NVIDIA Warp~\citep{macklin2022warp} instead of the built-in IsaacSim~\citep{NVIDIA_Isaac_Sim} renderer to achieve higher rendering throughput during training~\citep{zhang2026rpl}. The depth image resolution is $60\times80$.
To improve robustness, we apply minimal depth randomization consisting of Gaussian noise $\mathcal{N}(0,0.005)$ and pixel dropout with probability $0.05$. The resulting depth values are clipped to the range $[0.1\,\mathrm{m},\,2\,\mathrm{m}]$. Since the onboard camera of Unitree G1 is not rigidly mounted due to hardware constraints, we additionally randomize camera extrinsics with translational perturbations in $[-0.01\,\mathrm{m},\,0.01\,\mathrm{m}]$ and rotational perturbations in $[-1.5^\circ,\,1.5^\circ]$.

The actor and critic networks use MLP architectures of
$[2048,1024,512,256,128]$ and $[512,256,128]$, respectively. The depth encoder consists of a three-layer convolutional network with channel dimensions $[8,8,8]$ and kernel size $5$.

\subsection{Bridging the Sim-to-Real Gap}
\label{appendix:method3}

The unified visuomotor climbing policy faces two primary sources of sim-to-real discrepancy: perception and dynamics. To mitigate the perception gap, we apply depth randomization during training and further leverage FastFoundationStereo~\citep{wen2026fastfoundationstereo} to obtain cleaner and more geometrically consistent depth observations in the real world.
On the dynamics side, we identify collision geometry as a critical yet often overlooked factor for successful sim-to-real transfer. This effect is particularly pronounced for ladder climbing, which involves frequent whole-body contacts and requires precise interaction between the robot and ladder.

\begin{figure}[!t]
\vspace{-10pt}
\centering
\includegraphics[width=\linewidth]{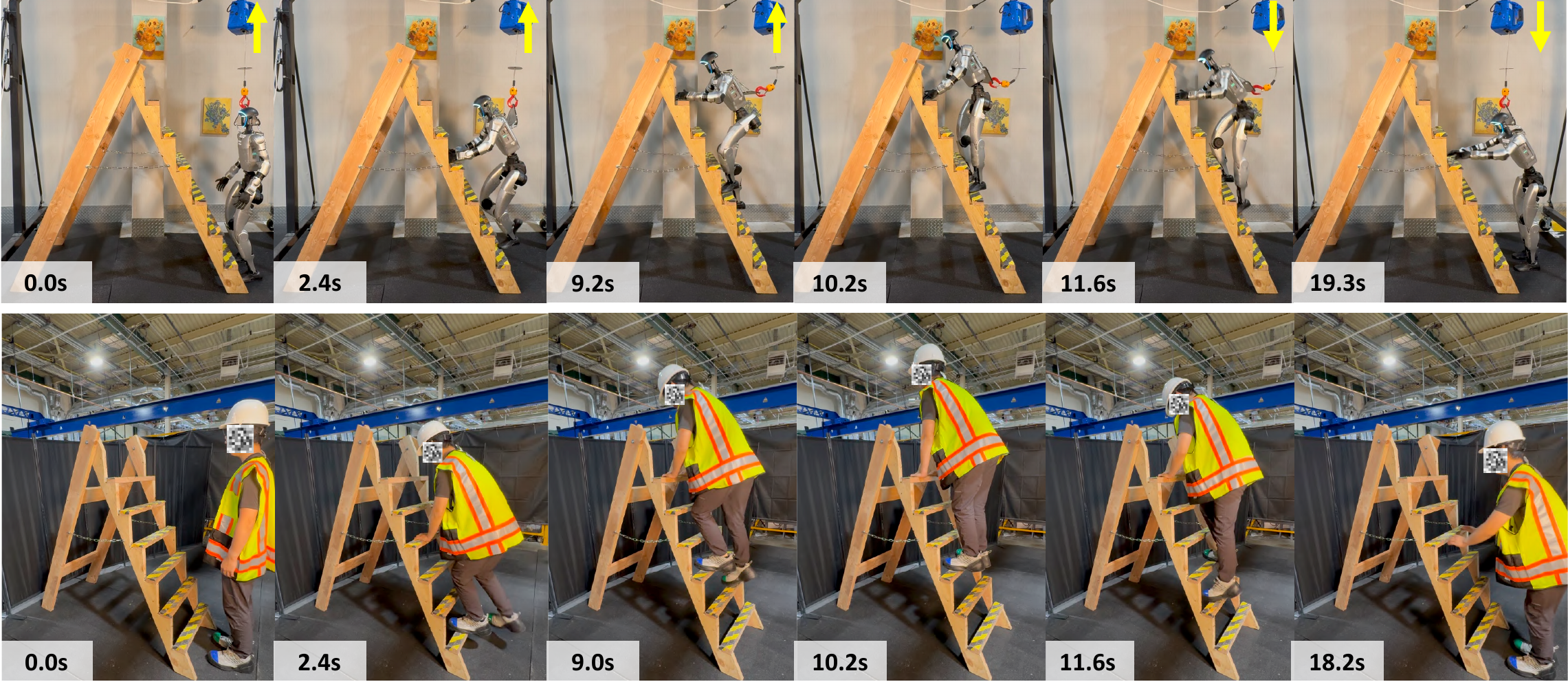}
\caption{Side-by-side comparison between a human and LadderMan performing the same ladder climbing task. LadderMan achieves human-comparable climbing speed.}
\vspace{-6pt}
\label{fig:appendix_human}
\end{figure}

\begin{wrapfigure}{r}{0.55\textwidth}
\vspace{-10pt}
\centering
\includegraphics[width=1.\linewidth]{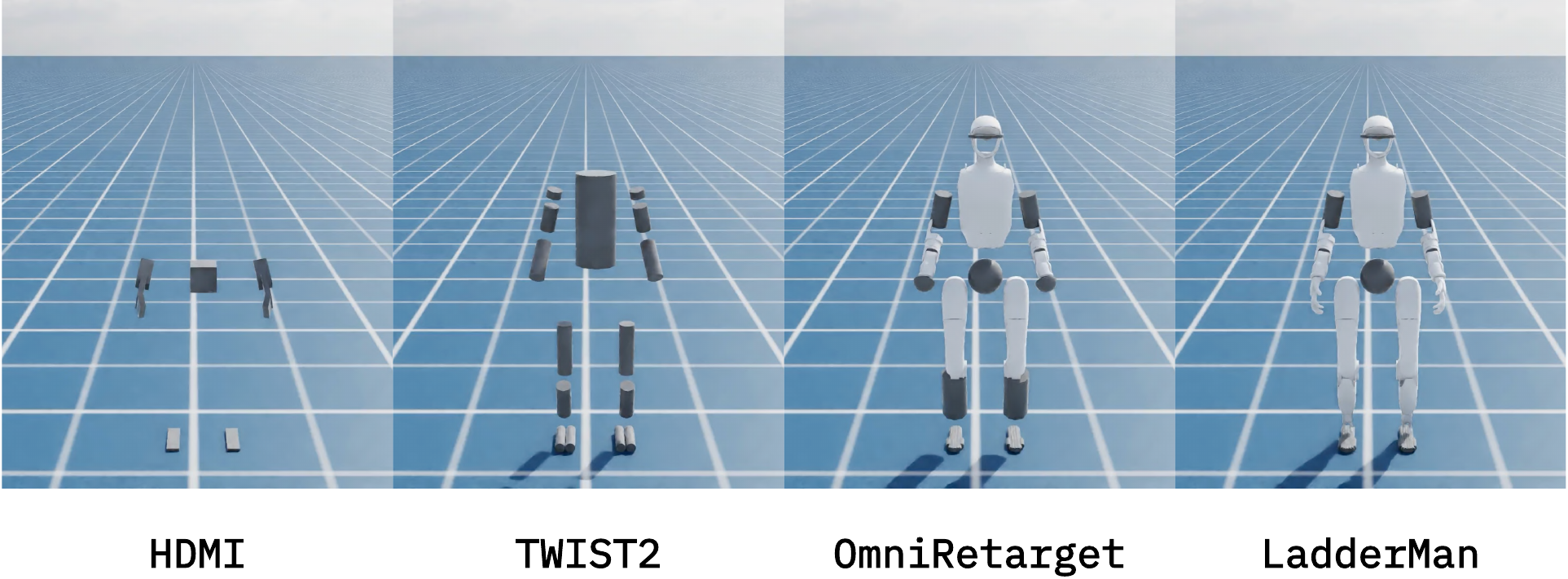}
\caption{Comparison of Unitree G1 collision geometries used in different humanoid projects.
}
\vspace{-10pt}
\label{fig:appendix_collision}
\end{wrapfigure}
\textbf{Collision Geometry.}
We investigate several publicly available Unitree G1 models used in recent humanoid projects and observe substantial differences in their collision geometry. In particular, as shown in Fig.~\ref{fig:appendix_collision}, the collision models used by TWIST2~\citep{ze2025twist2} and HDMI~\citep{weng2025hdmi} are heavily simplified, making accurate ladder contact simulation difficult.
We therefore adopt the collision geometry from OmniRetarget~\citep{yang2025omniretargetinteractionpreservingdatageneration} as our starting point due to its substantially higher geometric fidelity. However, we further observe that several body parts, particularly the shins, hands, and feet, are still approximated using simple collision primitives. While such approximations are often sufficient for locomotion on flat terrain, they introduce noticeable sim-to-real discrepancies for ladder climbing, where small contact errors can accumulate and lead to failure.
These observations highlight the importance of high-fidelity collision geometry for contact-rich humanoid tasks and motivate the collision model used in LadderMan.

\subsection{On-Ladder Manipulation via Dual-Agent Learning}
\label{appendix:method4}

In this stage, our goal is to train a manipulation policy
$
\pi^{\text{manip}}(s_t,g_t)
=
\left[
\pi^{u}(s_t,g_t);
\pi^{l}(s_t)
\right],
$
using a dual-agent formulation that maintains stable balance on the ladder while tracking teleoperated manipulation targets.
The two agents use separate actor and critic networks. The actor and critic networks use MLP architectures of $[512,256,256]$ and $[512,256,128]$, respectively.

\section{Experimental Results}\label{appendix:exp}
\subsection{Experiment Setup}\label{appendix:exp_setup}
\textbf{Climbing and Manipulation Policy.}
Both policies are deployed on the onboard NVIDIA Jetson Orin computer with a control frequency of $50\,\mathrm{Hz}$.
To generate the teleoperation targets $g_t$ during deployment, we use a PICO 4 Ultra VR headset. The system provides whole-body motion streaming through a head-mounted display, handheld controllers, and two ankle-mounted trackers, without requiring external motion capture systems.

\textbf{Vision Foundation Model.}
To obtain cleaner depth observations in the real world, we use Fast FoundationStereo~\citep{wen2026fastfoundationstereo} to process stereo grayscale images captured by the onboard Intel RealSense D435i camera. During data acquisition, the infrared emitter and automatic exposure control are disabled. Real-time inference is performed on an NVIDIA RTX 4090 GPU during deployment.

\begin{figure}[!t]
\centering
\vspace{-10pt}
\includegraphics[width=\linewidth]{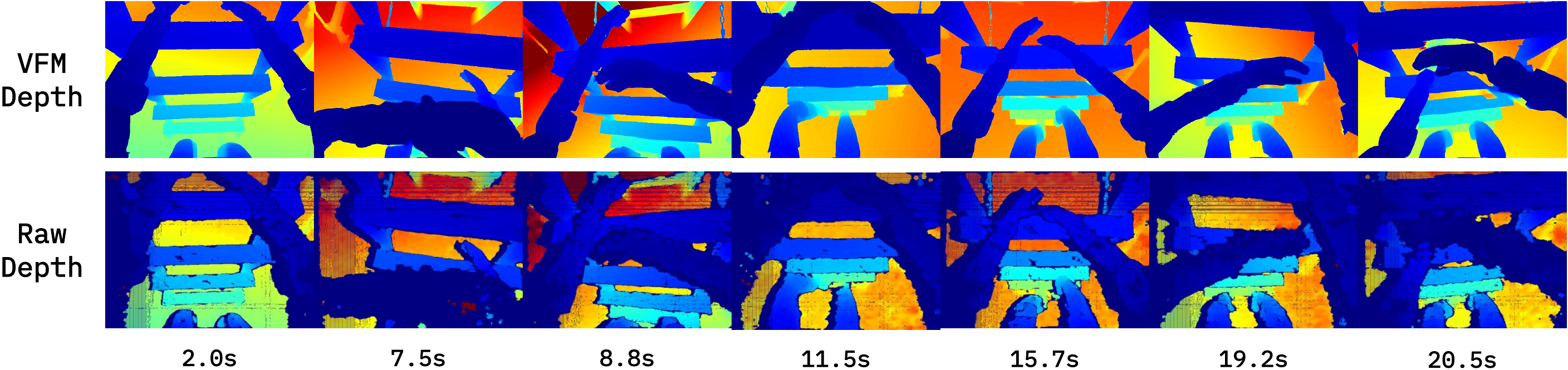}
\caption{Comparison between raw depth observations and depth predictions generated by Fast FoundationStereo. The foundation model significantly reduces sensor artifacts and improves geometric consistency, facilitating sim-to-real transfer.}
\vspace{-6pt}
\label{fig:appendix_depth}
\end{figure}

\subsection{Additional Results}\label{appendix:exp_results}
\begin{wrapfigure}{r}{0.4\textwidth}
\vspace{-10pt}
\centering
\includegraphics[width=1.\linewidth]{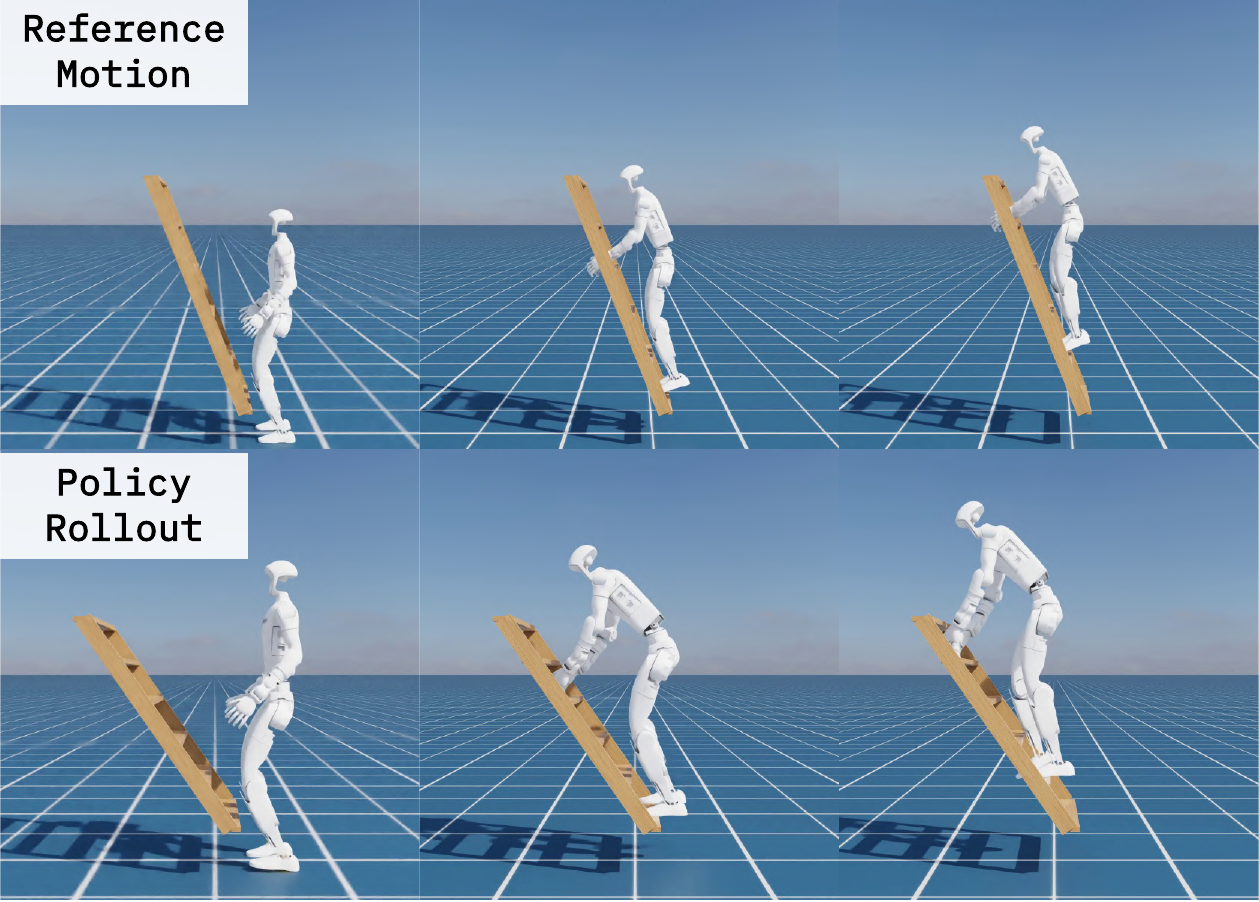}
\caption{Comparison between the reference motion and the learned climbing behavior on a ladder different from that used during motion capture.
}
\vspace{-10pt}
\label{fig:appendix_reference}
\end{wrapfigure}
\textbf{Hybrid Motion Tracking.}
We provide a side-by-side comparison between the collected reference motion and the learned policy execution on a ladder configuration with $(\theta,s)=(55^\circ,20\,\mathrm{cm})$, as shown in Fig.~\ref{fig:appendix_reference}.
Despite being trained from a single reference motion collected on a different ladder configuration, the learned policy successfully adapts the contact locations of both hands and feet while preserving the overall climbing pattern. This result qualitatively demonstrates the ability of hybrid motion tracking to generalize across ladder geometries without requiring additional reference motions.

\textbf{Human Comparison.}
As shown in Fig.~\ref{fig:appendix_human}, we provide a side-by-side comparison between a human and the humanoid robot performing the same ladder climbing task. For a complete traversal consisting of climbing three rungs up and three rungs down, the robot requires approximately $19.3\,\mathrm{s}$, while the human completes the task in approximately $18.2\,\mathrm{s}$.
Although the robot adopts a more conservative climbing strategy for safety and robustness, its climbing speed is already comparable to that of a human operator. This result suggests that LadderMan achieves practical climbing efficiency while maintaining reliable whole-body coordination.

\textbf{Depth Visualization.}
As shown in Fig.~\ref{fig:appendix_depth}, we provide side-by-side visualizations of the raw depth observations and the corresponding depth predictions generated by Fast FoundationStereo~\citep{wen2026fastfoundationstereo}.
Compared with raw depth observations, the predicted depth maps exhibit significantly reduced sensor artifacts and improved geometric consistency, particularly around ladder rungs and contact regions. This qualitative result supports our observation that vision foundation models reduce the need for extensive manually engineered depth randomization during sim-to-real transfer.

\end{document}